\documentclass{article}
\usepackage{iclr2027_conference,times}

\usepackage{amsmath,amsfonts,bm}

\def\eqref#1{equation~\ref{#1}}
\def\1{\bm{1}}

\def\va{{\bm{a}}}
\def\vb{{\bm{b}}}

\def\ve{{\bm{e}}}

\def\vg{{\bm{g}}}

\def\vp{{\bm{p}}}

\def\vr{{\bm{r}}}

\def\vu{{\bm{u}}}
\def\vv{{\bm{v}}}

\def\vy{{\bm{y}}}

\DeclareMathAlphabet{\mathsfit}{\encodingdefault}{\sfdefault}{m}{sl}
\SetMathAlphabet{\mathsfit}{bold}{\encodingdefault}{\sfdefault}{bx}{n}

\usepackage{amsmath,amssymb}
\usepackage{booktabs}
\usepackage{graphicx}
\usepackage{microtype}
\usepackage{xcolor}

\usepackage[utf8]{inputenc} % allow utf-8 input
\usepackage[T1]{fontenc}    % use 8-bit T1 fonts
\definecolor{Highlight}{rgb}{0.12,0.49,0.85}
\PassOptionsToPackage{hyphens}{url}
\usepackage[breaklinks,colorlinks,linkcolor=purple,citecolor=Highlight]{hyperref}
\usepackage{url}            % simple URL typesetting
\usepackage{amsfonts}       % blackboard math symbols
\usepackage{nicefrac}       % compact symbols for 1/2, etc.
\usepackage{pifont}
\usepackage{enumitem}
\usepackage{multirow}
\usepackage{longtable}
\usepackage{needspace}

\title{When Tomorrow Becomes Today: Self-Evolving Policies for Agentic Time-Series Forecasting}

\author{Yifan Hu$^{1,2,}$\thanks{ Equal contribution \ \ $^{\dagger}$ Corresponding author}, Xilin Dai$^{1,3,*}$, Zhiyuan Qu$^{2}$, Yiding Liu$^{1}$, Zewei Dong$^{1,\dagger}$, Jiang-ming Yang$^{1}$, Qiang Xu$^{3,\dagger}$ \\
$^{1}$Ant International $^{2}$Tsinghua University 
$^{3}$The Chinese University of Hong Kong \\
\texttt{\{hyf476357\}@ant-intl.com}
}

\newcommand{\method}{\textsc{TimEvolve}}
\newcommand{\methodstack}{\textsc{TimEvolve-Stack}}
\newcommand{\pack}{\textsc{Forecast-Pack}}

\newcommand{\info}{\mathcal{I}}
\newcommand{\policy}{\boldsymbol{\Pi}}
\newcommand{\memory}{\mathcal{M}}
\newcommand{\clip}{\operatorname{clip}}

\usepackage[capitalize]{cleveref}
\crefname{section}{Sec.}{Secs.}
\crefname{section}{Section}{Sections}
\crefname{table}{Table}{Tables}
\crefname{table}{Tab.}{Tabs.}

\newcommand{\appref}[1]{App.~\ref{#1}}

\usepackage{threeparttable}
\usepackage{multirow, multicol}
\usepackage{graphicx}
\usepackage{caption}
\usepackage{subfigure}

\usepackage{amssymb}% http://ctan.org/pkg/amssymb
\usepackage{pifont}% http://ctan.org/pkg/pifont
\iclrfinalcopy
\begin{document}

\maketitle

\begin{abstract}
Agentic time series forecasting concerns systems whose underlying mechanisms evolve, making the relative effectiveness of numerical models, reasoning strategies, and intervention rules inherently time-varying. Consequently, a time series agent must adapt the forecasts it produces and the orchestration policy that determines which components to trust and how to coordinate them. 
The deployment process naturally provides supervision for this adaptation as forecast horizons elapse and realized targets reveal the effectiveness of earlier decisions. Committing all numerical expert forecasts and candidate agent paths before target observation allows each realized outcome to evaluate the entire alternative set, providing delayed feedback without additional annotation. 
However, existing time series agents primarily incorporate prior experience through forecast refinement, reflection, or retrieval, without systematically converting realized outcomes into persistent updates to the joint orchestration policy governing later origins.
To exploit this delayed feedback systematically, we introduce \method{}, a frozen-backbone time series agent that converts each realized outcome into persistent joint updates of expert trust, agent path selection, and intervention strength. A temporally ordered predict, reveal, and update protocol applies this feedback to subsequent forecasts.
Experiments across eight Time-MMD domains show that \method{} achieves the best average MSE and MAE ranks among fifteen methods and the lowest errors on both metrics in seven domains.
These results demonstrate the value of learning forecasting policies from the futures encountered during deployment.
\end{abstract}

\section{Introduction}

Agentic time series forecasting addresses systems whose underlying mechanisms evolve as trends drift, regimes shift, and external events alter system dynamics~\citep{chang2026a,zhao2025timeseriesscientist}. Consequently, the relative effectiveness of numerical models, reasoning strategies, and intervention rules is inherently time-varying, and a component that supports accurate forecasting under one set of temporal conditions can become unreliable under another~\citep{cheng2026agentic,kim2025comprehensive}. Therefore, a time series agent must adapt not only the forecasts it produces, but also the orchestration policy that determines which components to trust and how to coordinate them~\citep{hulandscape}.

The deployment process naturally provides supervision for this adaptation. As forecast horizons elapse, realized targets reveal the errors of earlier forecasts, supplying delayed feedback for online learning~\citep{joulani2013online,zhang2023onenet}. In sequential decision-making, evaluating alternative policies from logged interaction trajectories is an off-policy estimation problem~\citep{jiang2016doubly}. Non-performative forecasting instead permits direct comparison of precommitted forecasts against a shared outcome, following the full-information setting of prediction with expert advice~\citep{cesabianchi2006prediction}. We extend this feedback structure to numerical experts and candidate agent paths fixed before reveal, where each \emph{agent path} is an evidence-grounded plan specifying expert selection and optional corrections to produce a candidate forecast.
Each reveal provides task-native supervision for both the selected and unselected alternatives, enabling prior forecasting decisions to be evaluated directly from observed outcomes. When tomorrow becomes today, the environment evaluates both what the agent predicted and how that prediction was constructed.

\begin{figure*}[t]
    \centering
    \includegraphics[width=0.98\textwidth]{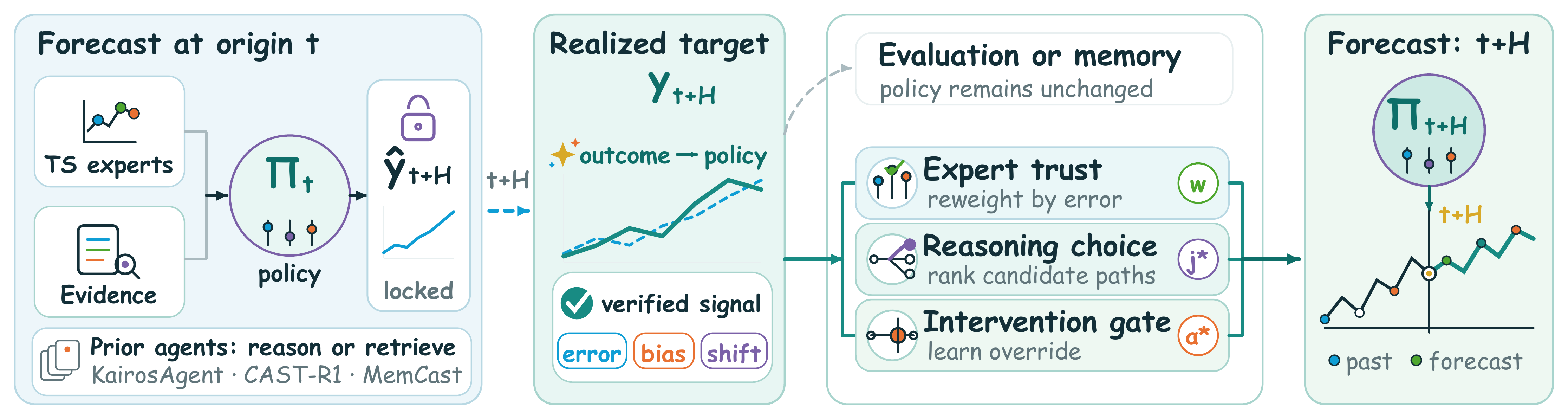}
    \caption{When outcomes become supervision, forecast evaluation gives way to policy evolution.
    Before observing the target, a time series agent commits its forecast and decision trace.
    Evaluation or passive memory leaves the orchestration policy unchanged, whereas \method{} uses the realized outcome to update expert trust, agent path selection, and intervention strength for subsequent forecasts.
    }
    \label{fig:motivation}
\end{figure*}

Recent work has expanded time series analysis and forecasting into agentic workflows~\citep{cheng2026agentic} involving tool use~\citep{wu2026timeart,tao2026anomamind}, learned sequential decision policies~\citep{tao2026castr1}, semantic reasoning~\citep{zhou2026veritime,guan2025timeomni}, specialized numerical forecasters~\citep{feng2026kairosagent}, iterative forecast refinement~\citep{liao2026lastmile}, reflection~\citep{wang2024news}, and memory~\citep{tao2026memcast,lyu2026tsmemory}.
As shown in \cref{fig:motivation}, these systems improve current forecast construction and incorporate prior experience through refinement, reflection, or retrieval. However, they do not systematically convert realized deployment outcomes into persistent updates to an explicit joint orchestration policy governing later forecasting decisions. \textit{This creates a mismatch between the time-varying effectiveness of individual components and the comparatively static rule used to coordinate models, evidence, reasoning, and intervention.} Revealed errors can thus inform retrieval or local forecast refinement without directly recalibrating the persistent decision rules used in subsequent forecasts.

Addressing this orchestration gap requires adapting three coupled decisions. 
\ding{182} The agent must determine which numerical experts to trust, since their relative reliability changes as temporal conditions evolve. 
\ding{183} It must select an evidence-grounded agent path, because alternative interpretations of shared numerical and contextual evidence can favor different expert subsets and corrections. 
\ding{184} It must determine how strongly the selected path should modify the numerical prior, balancing useful corrections against errors introduced by weak or misinterpreted evidence. 
These decisions are \textit{interdependent} because expert trust influences both the numerical prior and the candidate forecasts, path selection determines the proposed correction relative to the prior, while intervention strength controls the extent to which this correction is included in the final forecast.  
Revising expert trust can therefore change which path is most useful and how strongly it should be adopted, while selecting a different path changes the required calibration of the correction. 
These dependencies necessitate a joint orchestration policy that uses the same realized outcome to inform coordinated updates to expert trust, agent path selection, and intervention strength, allowing subsequent decisions to adapt to changes in the reliability of both components and their interactions.

To realize this joint update, we introduce \method{}, a frozen-backbone time series agent with three complementary modules that convert delayed outcome feedback into persistent updates to the corresponding decision rules.
EvolveTrust updates the contribution of numerical experts to the time series prior, EvolveReason updates the preference over evidence-grounded agent paths, and EvolveIntervene updates how strongly the selected path modifies the numerical prior. Before observing the target, \method{} commits all numerical expert forecasts, candidate paths, selected decisions, and the final forecast. 
Once the forecast horizon elapses, the realized target provides expert errors, candidate forecast errors, and a target for calibrating the selected path's influence on the numerical prior.
The predict, reveal, and update protocol applies these updates to subsequent eligible forecasts. The language model, numerical forecasters, analytical tools, and prompts remain frozen while the orchestration policy evolves. 
Across eight Time-MMD domains, \method{} achieves the best average MSE and MAE ranks against agentic systems, foundation models, and full-shot forecasting models.
Our contributions are as follows.
\begin{itemize}[leftmargin=*,itemsep=-0.1em]
    \item \textbf{Delayed-Feedback Learning.} We formulate deployment forecasting as policy learning from delayed outcome feedback and develop a temporally valid predict, reveal, and update protocol. Each realized target scores the precommitted alternatives, and the resulting policy updates govern subsequent forecasts.
    \item \textbf{Policy Evolution.} We introduce \method{}, which converts realized outcome feedback into persistent, coordinated updates to expert trust, agent path selection, and intervention strength while keeping the language model, numerical forecasters, analytical tools, and prompts frozen.
    \item \textbf{Empirical Validation.} Extensive experiments on eight Time-MMD domains show that \method{} achieves the best average rank under both MSE and MAE and improves over its fixed-policy counterpart on both metrics in every domain.
\end{itemize}

\section{Related Work}

\subsection{Agentic time series forecasting}

Agentic time series forecasting extends model-centric prediction into a workflow that coordinates numerical forecasters, analytical tools, contextual evidence, semantic reasoning, iterative refinement, reflection, and memory~\citep{xia2026into, cheng2026agentic}. Cast-R1 learns tool-augmented sequential decision policies through supervised and reinforcement learning~\citep{tao2026castr1}. KairosAgent integrates tool-grounded semantic reasoning with a time series foundation model and optimizes its reasoner using forecasting-oriented objectives~\citep{feng2026kairosagent}. Last-mile agents use weakly structured contextual evidence to revise numerical forecasts and retain post-hoc reflections for later use~\citep{liao2026lastmile}. These methods broaden time series forecasting from a numerical mapping into an agentic process that selects tools, interprets evidence, invokes specialized forecasters, and refines candidate predictions.
These approaches develop forecast construction and experience use. \method{} makes the persistent orchestration policy the unit of adaptation, using realized outcomes to update expert trust, agent path selection, and intervention strength across deployment.

\subsection{Outcome-driven adaptation}

Learning from revealed outcomes is central to online prediction with expert advice, where observed losses update the relative influence of competing experts~\citep{freund1997decision,cesabianchi2006prediction}. OneNet addresses concept drift in time series forecasting through online ensembling~\citep{zhang2023onenet}. MoE-F develops online gating for mixtures of language model experts in time series prediction~\citep{saqur2025filtered}. These methods use outcome feedback to adapt expert allocation.
A complementary line of work incorporates feedback through agent memory. Reflexion converts environmental feedback into textual experience that guides subsequent attempts~\citep{shinn2023reflexion}. In time series forecasting, MemCast constructs hierarchical experience memory and adapts memory-entry confidence during inference, allowing retrieved experience to influence trajectory selection and reflection~\citep{tao2026memcast}. \method{} combines these adaptation targets in a joint orchestration policy. Each realized outcome scores the precommitted numerical expert forecasts and candidate agent paths and provides an intervention target, supervising expert trust, agent path selection, and intervention strength together.

\section{Preliminary: Delayed-Feedback Policy Learning}
\label{sec:problem}

We consider a scalar target series $\{x_1,\ldots,x_T\}$ and a fixed forecast horizon $H$. At forecasting origin $t$, the history $x_{1:t}$ and time-aligned context $c_t$ are available, while the target window is
\begin{equation}
    \vy_t=[x_{t+1},\ldots,x_{t+H}]^\top\in\mathbb{R}^{H}.
    \label{eq:target-window}
\end{equation}
Both $\hat{\vy}_t$ and $\vy_t$ use the forecast issue time $t$ as their index. The information set $\info_t$ contains the available history and context, fixed predeployment calibration artifacts, and observed outcomes of windows completed by $t$. A forecast issued at $s$ supplies feedback when $s+H\leq t$. Overlapping forecasts remain outstanding until their corresponding target windows have elapsed.

At reveal, let $\Omega_s\subseteq\{1,\ldots,H\}$ contain the observed coordinates of $\vy_s$, and let $m_s=|\Omega_s|$. For $m_s>0$, the masked inner product and MSE are
\begin{equation}
    \langle\vu,\vv\rangle_{\Omega_s}=\sum_{h\in\Omega_s}u_hv_h,
    \qquad
    \operatorname{MSE}_{\Omega_s}(\vu,\vv)
    =\frac{\|\vu-\vv\|_{\Omega_s}^{2}}{m_s},
    \label{eq:masked-loss}
\end{equation}
where $\|\vu\|_{\Omega_s}^{2}=\langle\vu,\vu\rangle_{\Omega_s}$. These observed coordinates define all target-dependent losses; windows with $m_s=0$ leave the state unchanged. Pre-reveal features are computed from $\info_t$.

Let $\mathcal{K}_t\subseteq\{1,\ldots,K\}$ index the active numerical experts and $\mathcal{J}_t\subseteq\{1,\ldots,J\}$ the valid agent paths. They produce indexed forecast collections $\mathcal{F}_t=\{\hat{\vy}_t^{(k)}:k\in\mathcal{K}_t\}$ and $A_t=\{\va_t^{(j)}:j\in\mathcal{J}_t\}$, with every curve in $\mathbb{R}^{H}$. A path is a structured decision whose execution produces $\va_t^{(j)}$; its index $j$ is local to that origin. The learner state is
\begin{equation}
    \policy_t=\bigl(\policy_t^{\mathrm{trust}},
    \policy_t^{\mathrm{reason}},\policy_t^{\mathrm{intervene}}\bigr),
    \qquad
    \hat{\vy}_t=G(\info_t,\mathcal{F}_t,A_t;\policy_t).
    \label{eq:policy-state}
\end{equation}
Each component stores decision parameters and update statistics. In particular, $\policy_t^{\mathrm{reason}}$ contains the selector coefficients $\boldsymbol{\theta}_t$, and $\policy_t^{\mathrm{intervene}}$ contains the gate coefficients $\boldsymbol{\omega}_t$ shown in \cref{fig:method-pipeline}. The fixed map $G$ constructs and combines forecasts under this state.

Before observing $\vy_s$, the agent stores its candidates, features, and decisions in a record $\mathcal{R}_s$. At $s+H$, its feedback updates the latest active state,
\begin{equation}
    \policy_{s+H}
    =U\!\left(\policy_{s+H}^{-},\mathcal{R}_s,\vy_s,\Omega_s\right).
    \label{eq:delayed-update}
\end{equation}
The state $\policy_{s+H}^{-}$ includes updates from other completed windows and immediately precedes this reveal. The learner processes feedback before issuing the forecast at $s+H$, with earlier forecasts already committed. Targets are unaffected by forecast selection, so all stored alternatives share the same observed outcome.

\section{Method}
\label{sec:method}
\suppressfloats[t]

\subsection{Overview}

\begin{figure*}[t]
    \centering
    \includegraphics[width=\textwidth]{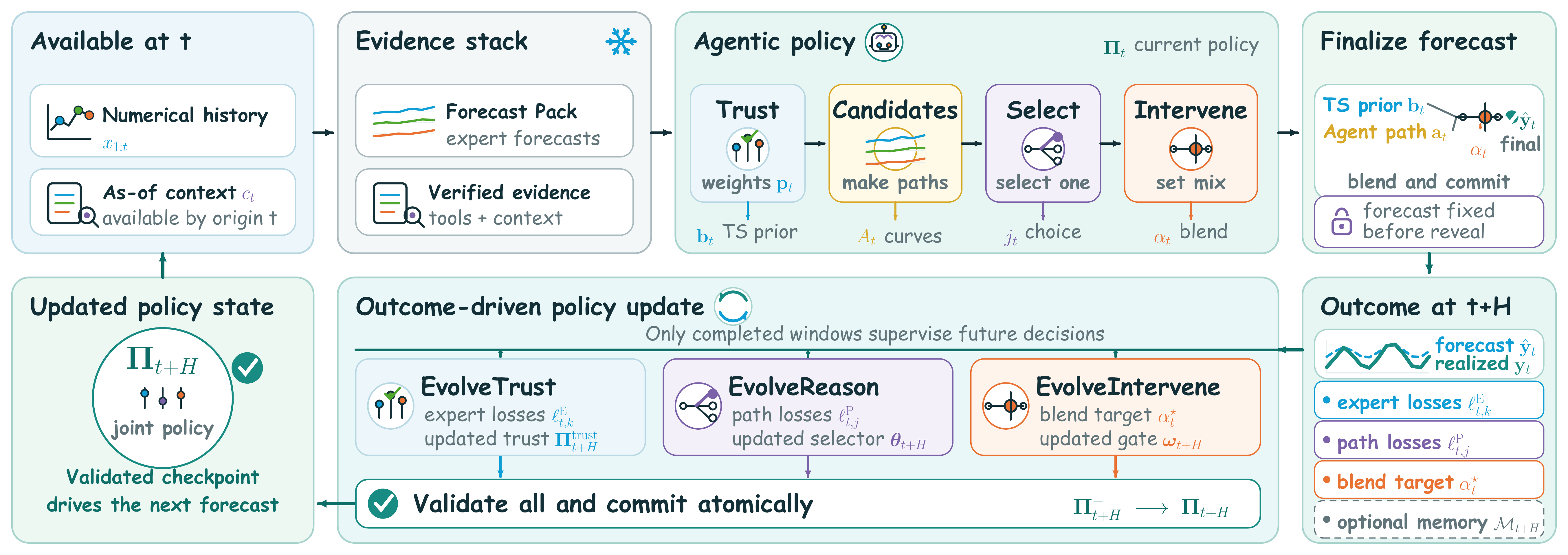}
    \caption{\textbf{\method{} predict, reveal, and update pipeline.} History $x_{1:t}$ and context $c_t$ produce expert forecasts and the candidate set $A_t$. Expert weights $\vp_t$ form the prior $\vb_t$, selection $j_t$ yields $\va_t=\va_t^{(j_t)}$, and the gate commits $\hat{\vy}_t=(1-\alpha_t)\vb_t+\alpha_t\va_t$. At $t+H$, observed coordinates of $\vy_t$ yield expert losses $\ell^{\mathrm{E}}_{t,k}$, path losses $\ell^{\mathrm{P}}_{t,j}$, and the blend target $\alpha_t^\star$. They update the latest state $\policy_{t+H}^{-}$ to $\policy_{t+H}$, including expert trust, selector coefficients $\boldsymbol{\theta}_{t+H}$, and gate coefficients $\boldsymbol{\omega}_{t+H}$. Episodic memory $\memory_{t+H}$ is maintained as auxiliary context.}
    \label{fig:method-pipeline}
\end{figure*}

\method{} implements the loop in \cref{fig:method-pipeline} through three linked decisions. EvolveTrust forms expert weights $\vp_t$ and the numerical prior $\vb_t$. A frozen language model proposes structured paths, whose execution under these weights produces $A_t$. EvolveReason selects $j_t$ and hence $\va_t=\va_t^{(j_t)}$, and EvolveIntervene chooses its influence $\alpha_t$ on the prior. Expert trust therefore affects both the prior and the candidates, while the selected path determines the correction calibrated by the gate. Revealed outcomes update these linked decision rules while the forecasting components remain frozen.

\subsection{Evidence-grounded forecast construction}

\paragraph{Forecasts and evidence.}
The \pack{} supports frozen time series foundation models and statistical filters for level, trend, seasonality, and changing dynamics. The main evaluation uses a foundation-model-augmented numerical tool pool, with the same frozen forecasters shared across \method{} and its policy ablations (\appref{app:run-scope}). Active experts produce aligned $H$-step forecasts from the available history. For each expert, $\ve_{t,k}$ summarizes past reliability, current forecast plausibility, and agreement with other sources. Fixed analytical tools extract temporal structure from the history, while $c_t$ supplies reports and search evidence available by $t$. Numerical diagnostics characterize each forecast relative to the observed temporal pattern, while contextual evidence informs expert selection and local corrections. \appref{app:method-details} details these diagnostics.

\paragraph{Structured paths.}
Each proposed path specifies an evidence interpretation, an expert subset $S_t^{(j)}\subseteq\mathcal{K}_t$, and an optional correction $\boldsymbol{\Delta}_t^{(j)}\in\mathbb{R}^{H}$. Current trust weights define the initial mixture
\begin{equation}
    u_{t,k}^{(j)}=
    \frac{p_{t,k}\mathbb{1}[k\in S_t^{(j)}]}
         {\sum_{q\in S_t^{(j)}}p_{t,q}}.
    \label{eq:path-weights}
\end{equation}
Valid paths select nonempty subsets with positive trust mass. The configured source-budget projection yields normalized weights $\tilde u_{t,k}^{(j)}$ supported on the selected subset, limiting the influence of correlated expert families. The runtime then constructs
\begin{equation}
    \va_t^{(j)}=\sum_{k\in\mathcal{K}_t}
    \tilde u_{t,k}^{(j)}\hat{\vy}_t^{(k)}
    +\boldsymbol{\Delta}_t^{(j)}.
    \label{eq:path-execution}
\end{equation}
Paths with $\boldsymbol{\Delta}_t^{(j)}=0$ can still differ from the prior through their selected expert mixture. Nonzero corrections add evidence-linked adjustments to that mixture. Valid paths form the nonempty set $\mathcal{J}_t$. The language model interprets the evidence, and the runtime executes and stores the resulting candidate forecasts together with their structured decisions.

\subsection{EvolveTrust for Expert Trust}
\label{sec:trust}

EvolveTrust combines four views of expert reliability. The warm-up view uses training and validation forecasts, the online view uses completed deployment windows, the current-evidence view assesses forecast plausibility, and the pattern view summarizes performance under previously observed evidence patterns. Together, these views distinguish aggregate expert performance from reliability under different observed evidence patterns and current forecast plausibility. Each $\vp_t^v$ is a probability vector over active experts, with zero mass outside $\mathcal{K}_t$. Their mixture forms the prior,
\begin{equation}
    \vp_t=\sum_v\beta_{t,v}\vp_t^v,
    \quad \beta_{t,v}\geq0,\quad\sum_v\beta_{t,v}=1,
    \qquad
    \vb_t=\sum_{k\in\mathcal{K}_t}p_{t,k}\hat{\vy}_t^{(k)}.
    \label{eq:ts-prior}
\end{equation}
The view maps and mixture schedule are fixed calibration choices. The online and pattern contributions enter through $\rho_t=N_t/(N_t+\tau)$, where $N_t$ counts completed deployment windows with observed targets and $\tau>0$ controls cold start. The ramp increases their contribution as labeled windows accumulate, while warm-up and current evidence support prediction during cold start.

Let $\mathcal{C}_{t,k}$ contain deployment origins $s$ with $s+H\leq t$, $m_s>0$, and a stored expert-$k$ forecast. With per-window losses $\ell^{\mathrm{E}}_{s,k}=\operatorname{MSE}_{\Omega_s}(\hat{\vy}_s^{(k)},\vy_s)$, the online view uses inverse-error weighting,
\begin{equation}
    \operatorname{RMSE}_{t,k}
    =\left(\frac{1}{|\mathcal{C}_{t,k}|}
    \sum_{s\in\mathcal{C}_{t,k}}\ell^{\mathrm{E}}_{s,k}\right)^{1/2},
    \qquad
    p^{\mathrm{online}}_{t,k}\propto
    \frac{1}{\operatorname{RMSE}_{t,k}+\epsilon}.
    \label{eq:online-trust}
\end{equation}
Here $\epsilon>0$, and the error estimate applies to experts with eligible observations. Each reveal scores all stored experts, including those outside the selected path, and updates reliability and pattern statistics. Updated trust changes both the numerical prior and future path mixtures.

\subsection{EvolveReason for Agent Path Selection}
\label{sec:reason}

EvolveReason ranks candidates using five features available before reveal,
\begin{equation}
    \boldsymbol{\psi}_t^{(j)}=
    [1,\;\text{continuity},\;\text{in-range},\;\text{smoothness},
    \;\text{structural score}]^\top\in\mathbb{R}^{5},
    \label{eq:path-features}
\end{equation}
computed from $\va_t^{(j)}$ and $x_{1:t}$. This representation relates candidate shape and boundary behavior to the observed history, allowing one selector to compare paths built from different expert subsets. A shared rolling ridge model predicts log-transformed MSE and selects the lowest-scoring candidate,
\begin{equation}
    \hat z_t^{(j)}={\boldsymbol{\psi}_t^{(j)}}^\top\boldsymbol{\theta}_t,
    \qquad j_t=\arg\min_{j\in\mathcal{J}_t}\hat z_t^{(j)}.
    \label{eq:path-selection}
\end{equation}
During cold start, the fixed structural score selects the path. Subsequently, every observed target supplies $z_s^{(j)}=\log(1+\ell^{\mathrm{P}}_{s,j})$, where $\ell^{\mathrm{P}}_{s,j}=\operatorname{MSE}_{\Omega_s}(\va_s^{(j)},\vy_s)$. The monotone log transform preserves observed loss ordering while compressing large errors in the regression targets. The shared rolling fit in \appref{app:rolling-fits} pairs saved features from every valid candidate, including unselected ones, with the outcome-derived target for that candidate forecast.

\subsection{EvolveIntervene for Intervention Strength}
\label{sec:intervene}

\paragraph{Pre-reveal decision.}
Let $\boldsymbol{\delta}_t=\va_t-\vb_t$ be the selected path's deviation from the prior. This includes both expert reselection and any local correction. The final forecast is
\begin{equation}
    \hat{\vy}_t=\vb_t+\alpha_t\boldsymbol{\delta}_t,
    \qquad
    \alpha_t=\clip(\vg_t^\top\boldsymbol{\omega}_t,0,1).
    \label{eq:intervention}
\end{equation}
Here $\clip(z,l,u)=\min\{u,\max\{l,z\}\}$. The pre-reveal features $\vg_t\in\mathbb{R}^{d_g}$ contain an intercept, shift diagnostics, expert disagreement, source-weight concentration, and text availability. The endpoints recover the prior and selected candidate, while intermediate values interpolate between their forecasts.

\paragraph{Outcome-derived target.}
After reveal, set $\vr_t=\vy_t-\vb_t$ and $D_t=\|\boldsymbol{\delta}_t\|_{\Omega_t}^{2}$. For the stored prior and path, the scalar blend minimizes
\begin{equation}
    L_t(\alpha)=\operatorname{MSE}_{\Omega_t}
    (\vb_t+\alpha\boldsymbol{\delta}_t,\vy_t)
    =\frac{\|\vr_t-\alpha\boldsymbol{\delta}_t\|_{\Omega_t}^{2}}{m_t}.
    \label{eq:blend-risk}
\end{equation}
For $D_t>0$, differentiating this quadratic gives the unconstrained optimum $\tilde\alpha_t$, and projection onto the admissible interval gives the supervision target,
\begin{equation}
    \tilde\alpha_t=\frac{\langle\boldsymbol{\delta}_t,\vr_t\rangle_{\Omega_t}}{D_t},
    \qquad
    \alpha_t^\star=\arg\min_{\alpha\in[0,1]}L_t(\alpha)
    =\clip(\tilde\alpha_t,0,1).
    \label{eq:alpha-star}
\end{equation}
The numerator measures alignment between the proposed correction and the prior's realized residual, while $D_t$ normalizes correction magnitude. Nonpositive alignment yields $\alpha_t^\star=0$, and sufficiently strong positive alignment yields full adoption. The target measures how strongly the selected path should have modified its committed prior and supplies a regression label for subsequent gate updates.

\paragraph{Loss geometry and weighted learning.}
Completing the square yields
\begin{equation}
    L_t(\alpha)=C_t+w_t(\alpha-\tilde\alpha_t)^2,
    \qquad
    w_t=\frac{D_t}{m_t},\quad
    C_t=\frac{\|\vr_t\|_{\Omega_t}^{2}}{m_t}-w_t\tilde\alpha_t^2.
    \label{eq:blend-quadratic}
\end{equation}
The quadratic weight $w_t$ quantifies the effect of coefficient error on forecast MSE. For $\tilde\alpha_t\in[0,1]$, the excess loss is $w_t(\alpha-\alpha_t^\star)^2$. \appref{app:blend-derivation} derives the general excess-loss expression.

EvolveIntervene fits $\vg_s$ to $\alpha_s^\star$ by rolling weighted ridge regression with $\bar w_s=\clip(w_s,10^{-4},10^4)$. This energy-weighted supervised objective emphasizes identifiable differences between the prior and selected path. The fit retains windows with $D_s$ above numerical tolerance. The resulting coefficients $\boldsymbol{\omega}_t$ predict $\alpha_t$ from current pre-reveal features.

\subsection{Outcome feedback and coordinated updates}
\label{sec:reveal}

The record $\mathcal{R}_s$ stores expert forecasts, executed candidates, evidence, pre-reveal features, and the committed decisions $(\vp_s,\vb_s,j_s,\va_s,\alpha_s,\hat{\vy}_s)$. At $s+H$, the observed target supplies expert losses $\ell^{\mathrm{E}}_{s,k}$, path losses $\ell^{\mathrm{P}}_{s,j}$, and the blend target $\alpha_s^\star$. This produces a labeled example for each recorded expert and candidate, together with a blend label when the selected correction is identifiable. These targets jointly supervise expert reliability, path ranking, and intervention.

The learner applies these signals to the latest state in \cref{eq:delayed-update} and publishes the three validated updates as one checkpoint. Its updated selector and gate coefficients are $\boldsymbol{\theta}_{s+H}$ and $\boldsymbol{\omega}_{s+H}$, respectively. Optional episodic summaries are stored separately as $\memory_{s+H}$. The updated trust, selector, and gate propagate feedback through expert aggregation, candidate execution, and subsequent forecasts.

\section{Experiments}
\label{sec:experiments}

\subsection{Experimental setup}

We evaluate \method{} on the eight Time-MMD~\citep{liu2024timemmd} domains using chronological $70/10/20$ training, validation, and test splits and each domain's benchmark lookback and four forecast horizons. We report MSE and MAE in standardized target space, excluding missing target coordinates. Each horizon is evaluated separately and contributes equally to its domain-level result. 
We then rank all fifteen methods within each domain and average the ranks across domains. Test forecasts use stride one, and policy updates use only forecasts whose complete horizons have elapsed. \appref{app:timemmd-protocol} provides the domain-specific configurations and details.

We compare \method{} with fourteen baselines spanning four groups. \ding{172} Agentic baselines: KairosAgent~\citep{feng2026kairosagent}, MemCast~\citep{tao2026memcast}, and CastFlow~\citep{pan2026castflow}. 
\ding{173} Zero-shot foundation models: Aurora~\citep{wu2025aurora}, TimesFM-3~\citep{jain2026timesfm3}, Chronos-2~\citep{ansari2025chronos2}, Toto-2.0 (313M)~\citep{khwaja2026toto2}, Sundial~\citep{liu2025sundial} and Moirai-Large~\citep{woo2024moirai}. 
\ding{174} Full-shot multimodal models: T3Time~\citep{chowdhury2026t3time}, TimeCMA~\citep{liu2025timecma}, and CALF~\citep{liu2025calf}.
\ding{175} Full-shot unimodal models include PatchTST~\citep{nie2023patchtst} and DLinear~\citep{zeng2023dlinear}.

\begin{table}[!t]
\caption{Results on the eight Time-MMD domains. Each domain cell reports MSE/MAE metrics, averaged equally over the four forecasting horizons. Average rank reports the corresponding ranks averaged across the eight domains among fifteen methods. The best and second-best results are in \textbf{bold} and \underline{underline} respectively.}
\label{tab:complete-domain-results}
\centering
\scriptsize
\setlength{\tabcolsep}{2.2pt}
\renewcommand{\arraystretch}{1.02}
\resizebox{\textwidth}{!}{%
\begin{tabular}{@{}l*{8}{c}c@{}}
\toprule
Model
& Agriculture
& Climate
& Economy
& Energy
& Environment
& Security
& \shortstack{Social Good}
& Traffic
& \shortstack{Avg. rank} \\
\midrule
\multicolumn{10}{l}{\emph{Agentic systems}} \\
\method{} & \textbf{0.129}/\textbf{0.227} & \textbf{0.314}/\textbf{0.402} & \textbf{0.181}/\textbf{0.329} & \textbf{0.181}/\textbf{0.303} & \textbf{0.274}/\textbf{0.372} & \textbf{58.937}/\textbf{3.285} & 0.808/0.381 & \textbf{0.087}/\textbf{0.176} & \textbf{1.250}/\textbf{1.375} \\
KairosAgent & 0.194/0.282 & 0.863/0.739 & \underline{0.186}/\underline{0.335} & 0.217/0.330 & 0.378/0.435 & 76.658/4.340 & \textbf{0.769}/\textbf{0.376} & 0.151/0.231 & 5.188/5.125 \\
MemCast & 0.218/0.313 & 0.897/0.752 & 0.196/0.339 & 0.207/0.318 & 0.289/0.390 & 69.481/3.876 & \underline{0.797}/0.378 & 0.132/0.200 & 4.375/5.250 \\
CastFlow & 0.220/0.303 & 0.890/0.748 & 0.193/0.338 & 0.204/0.316 & 0.286/0.388 & 68.900/3.860 & 0.810/0.383 & \underline{0.130}/\underline{0.198} & \underline{4.000}/\underline{4.125} \\
\midrule
\multicolumn{10}{l}{\emph{Zero-shot foundation models}} \\
Aurora & 0.282/0.356 & 0.863/0.747 & 0.275/0.412 & 0.251/0.370 & \underline{0.276}/\underline{0.379} & 72.763/4.085 & 0.828/0.506 & 0.162/0.289 & 7.688/8.938 \\
TimesFM-3 & 0.195/0.270 & 0.555/0.516 & 0.428/0.482 & 0.334/0.384 & 0.326/0.391 & \underline{63.468}/3.822 & 0.834/0.457 & 0.414/0.469 & 8.125/8.375 \\
Chronos-2 & \underline{0.133}/\underline{0.233} & \underline{0.321}/\underline{0.417} & 0.264/0.408 & 0.209/0.316 & 0.340/0.397 & 64.695/\underline{3.818} & 1.124/0.497 & 0.315/0.431 & 7.375/6.688 \\
Toto-2.0 (313M) & 0.330/0.346 & 0.393/0.450 & 0.565/0.556 & \underline{0.196}/\underline{0.314} & 0.329/0.394 & 244.928/5.691 & 0.818/0.469 & 0.649/0.640 & 9.375/9.750 \\
Sundial & 0.327/0.366 & 0.920/0.765 & 0.216/0.348 & 0.234/0.337 & 0.379/0.443 & 83.403/4.836 & 0.819/\underline{0.377} & 0.228/0.292 & 9.375/8.375 \\
Moirai-Large & 0.239/0.306 & 0.982/0.792 & 0.198/0.345 & 0.261/0.347 & 0.412/0.446 & 74.249/4.129 & 0.868/0.391 & 0.186/0.263 & 8.750/7.750 \\
\midrule
\multicolumn{10}{l}{\emph{Full-shot multimodal models}} \\
T3Time & 0.229/0.303 & 1.206/0.894 & 0.239/0.384 & 0.266/0.378 & 0.489/0.507 & 72.113/4.070 & 0.998/0.432 & 0.289/0.368 & 10.750/9.813 \\
TimeCMA & 0.318/0.360 & 1.282/0.926 & 0.262/0.412 & 0.351/0.447 & 0.536/0.533 & 72.011/4.113 & 1.092/0.578 & 0.297/0.412 & 12.250/13.188 \\
CALF & 0.241/0.311 & 1.199/0.895 & 0.223/0.370 & 0.258/0.373 & 0.537/0.509 & 73.267/4.040 & 0.890/0.416 & 0.227/0.305 & 10.438/9.438 \\
\midrule
\multicolumn{10}{l}{\emph{Full-shot unimodal models}} \\
PatchTST & 0.248/0.308 & 1.176/0.891 & 0.223/0.380 & 0.243/0.353 & 0.496/0.513 & 76.105/4.445 & 0.959/0.475 & 0.209/0.316 & 10.188/10.750 \\
DLinear & 0.377/0.396 & 1.036/0.807 & 0.218/0.370 & 0.233/0.346 & 0.591/0.627 & 82.521/4.891 & 0.891/0.448 & 0.219/0.315 & 10.875/11.063 \\
\bottomrule
\end{tabular}}
\end{table}

\subsection{Main results}
\cref{tab:complete-domain-results} compares \method{} with agentic systems, foundation models, and full-shot forecasting models across eight Time-MMD domains. We highlight three findings.
\ding{172} \textbf{Consistent cross-domain performance.} \method{} achieves the best average MSE/MAE ranks of $1.250/1.375$ among fifteen methods, leading both metrics in seven of eight domains. Its aggregate advantage therefore reflects broad domain-level gains under both error criteria, rather than isolated improvements in a few favorable settings.
\ding{173} \textbf{Gains over agentic baselines.} \method{} outperforms CastFlow in all sixteen domain--metric comparisons and both MemCast and KairosAgent in fourteen each. These results establish that a frozen-backbone agent can outperform existing agentic systems with adaptation concentrated in the orchestration policy rather than the underlying forecasting components.
\ding{174} \textbf{Effective orchestration across heterogeneous domains.} The strongest standalone foundation model varies across domains, with Chronos-2 leading this group in Agriculture and Climate, Toto-2.0 in Energy, and Aurora in Environment. \method{} surpasses these domain-specific competitors with a shared orchestration mechanism. This contrast supports adaptive coordination as an alternative to relying on a single forecaster whose relative strength varies across domains.

\subsection{Effect of policy evolution}
\label{sec:policy-evolution-results}

We assess both the overall benefit of policy evolution and the contribution of each update in \cref{tab:ablation}. All variants share the numerical experts, evidence and context pipeline, candidate-generation configuration, initialization, forecast origins, and reveal schedule. \methodstack{} holds expert trust, agent path selection, and intervention strength at their initial checkpoints throughout testing. Each \emph{w/o} variant instead freezes only the named policy component while the other two continue to update. The forward computation is retained in every variant, isolating the contribution of outcome-driven updates. \appref{app:ablation-settings} distinguishes these update interventions from restrictions on feedback coverage.

\begin{table}[!t]
\caption{Effect of policy evolution. Each ablation freezes one policy component at its initial checkpoint while preserving its forward computation; \methodstack{} freezes all three. Cells report standardized-space MSE/MAE averaged equally over four horizons. Lower is better; bold marks the best values, including ties at the reported precision.}
\label{tab:ablation}
\centering
\scriptsize
\setlength{\tabcolsep}{2.4pt}
\renewcommand{\arraystretch}{1.02}
\resizebox{\textwidth}{!}{%
\begin{tabular}{@{}l*{8}{c}@{}}
\toprule
Model
& Agriculture
& Climate
& Economy
& Energy
& Environment
& Security
& \shortstack{Social Good}
& Traffic \\
\midrule
w/o EvolveTrust 
& 0.142/0.236 & 0.334/0.424 & 0.205/0.349 & 0.183/0.307 & 0.281/0.385 & 59.712/3.391 & 0.812/0.448 & 0.088/0.178 \\
w/o EvolveReason 
& 0.138/0.234 & 0.328/0.418 & 0.196/0.342 & 0.182/0.305 & 0.278/0.380 & 59.402/3.351 & 0.810/0.421 & 0.088/0.177 \\
w/o EvolveIntervene 
& 0.135/0.232 & 0.324/0.414 & 0.190/0.337 & 0.182/0.304 & 0.276/0.378 & 59.248/3.329 & 0.809/0.408 & 0.088/0.177 \\
\methodstack{} 
& 0.148/0.240 & 0.342/0.432 & 0.220/0.360 & 0.185/0.309 & 0.285/0.396 & 60.429/3.479 & 0.814/0.508 & 0.089/0.178 \\
\method{} 
& \textbf{0.129/0.227} & \textbf{0.314/0.402} & \textbf{0.181/0.329} & \textbf{0.181/0.303} & \textbf{0.274/0.372} & \textbf{58.937/3.285} & \textbf{0.808/0.381} & \textbf{0.087/0.176} \\
\bottomrule
\end{tabular}}
\end{table}

Compared with \methodstack{}, \method{} lowers both errors in every domain. Averaging domain-wise relative error reductions gives a $6.3\%$ reduction in MSE and a $7.6\%$ reduction in MAE. The largest MSE reduction is $17.7\%$ in Economy, from $0.220$ to $0.181$, while the largest MAE reduction is $25.0\%$ in Social Good, from $0.508$ to $0.381$. Since both variants use the same frozen forecasters and forecast-construction pipeline, these improvements isolate the benefit of adapting orchestration rather than expanding the numerical tool pool.

The complete policy achieves lower error in all sixteen domain--metric comparisons against each single-component freezing variant. The average degradation is largest when EvolveTrust is frozen, followed by EvolveReason and EvolveIntervene. In Economy, freezing trust, path selection, and intervention increases MSE by $13.3\%$, $8.3\%$, and $5.0\%$, respectively. Expert reliability adaptation thus has the largest observed effect, while learned path selection and intervention each improve the use of the resulting numerical evidence. Each contrast measures the contribution of one update with the other two active, rather than an additive decomposition of the total gain.

\subsection{Value of feedback coverage}
\label{sec:feedback-coverage-results}

We isolate the value of supervising unselected alternatives by varying expert and path feedback coverage. All variants share the forecast-construction configuration and reveal schedule. \emph{All-alternative feedback} updates expert trust from every committed expert forecast and trains path selection on every valid candidate. \emph{Selected-path feedback} restricts only path supervision to the chosen path, whereas \emph{selected-expert feedback} restricts only expert supervision to the experts used by that path. \emph{Selected-only feedback} applies both restrictions. Every updating variant retains the intervention-learning rule and computes its blend target from its stored prior and selected path. The frozen-policy reference applies no outcome-driven updates.

\begin{table}[!t]
\caption{Effect of feedback coverage. Rows vary whether expert and path losses supervise all precommitted alternatives or only those used by the selected path. Cells report standardized-space MSE/MAE averaged equally over four horizons. All updating variants retain intervention learning. Lower is better; bold marks the best values.}
\label{tab:feedback-coverage-domains}
\centering
\small
\setlength{\tabcolsep}{4.4pt}
\resizebox{\textwidth}{!}{%
\begin{tabular}{@{}l*{8}{c}@{}}
\toprule
Feedback regime
& Agriculture
& Climate
& Economy
& Energy
& Environment
& Security
& \shortstack{Social Good}
& Traffic \\
\midrule
Selected-path feedback 
& 0.134/0.231 & 0.321/0.409 & 0.187/0.334 & 0.182/0.305 & 0.277/0.379 & 59.374/3.348 & 0.810/0.414 & 0.088/0.177 \\
Selected-expert feedback 
& 0.140/0.236 & 0.331/0.419 & 0.198/0.344 & 0.184/0.308 & 0.282/0.386 & 59.756/3.401 & 0.812/0.451 & 0.088/0.178 \\
Selected-only feedback 
& 0.145/0.239 & 0.336/0.424 & 0.208/0.352 & 0.184/0.308 & 0.283/0.392 & 60.158/3.448 & 0.813/0.485 & 0.089/0.178 \\
Frozen policy 
& 0.148/0.240 & 0.342/0.432 & 0.220/0.360 & 0.185/0.309 & 0.285/0.396 & 60.429/3.479 & 0.814/0.508 & 0.089/0.178 \\
All-alternative feedback 
& \textbf{0.129/0.227} & \textbf{0.314/0.402} & \textbf{0.181/0.329} & \textbf{0.181/0.303} & \textbf{0.274/0.372} & \textbf{58.937/3.285} & \textbf{0.808/0.381} & \textbf{0.087/0.176} \\
\bottomrule
\end{tabular}}
\end{table}

\cref{tab:feedback-coverage-domains} shows that supervising all precommitted alternatives improves all sixteen comparisons over selected-only feedback, with average domain-wise relative reductions of $5.0\%$ in MSE and $6.3\%$ in MAE. Selected-only feedback improves fourteen comparisons over the frozen policy and ties both Traffic metrics at the reported precision. Restricting expert supervision incurs at least as much error as restricting path supervision in every cell. These results support using revealed outcomes to update the policy and retaining loss feedback for unselected alternatives, with broader expert coverage providing the larger observed benefit.

\begin{figure*}[t]
    \centering
    \includegraphics[width=\textwidth]
    {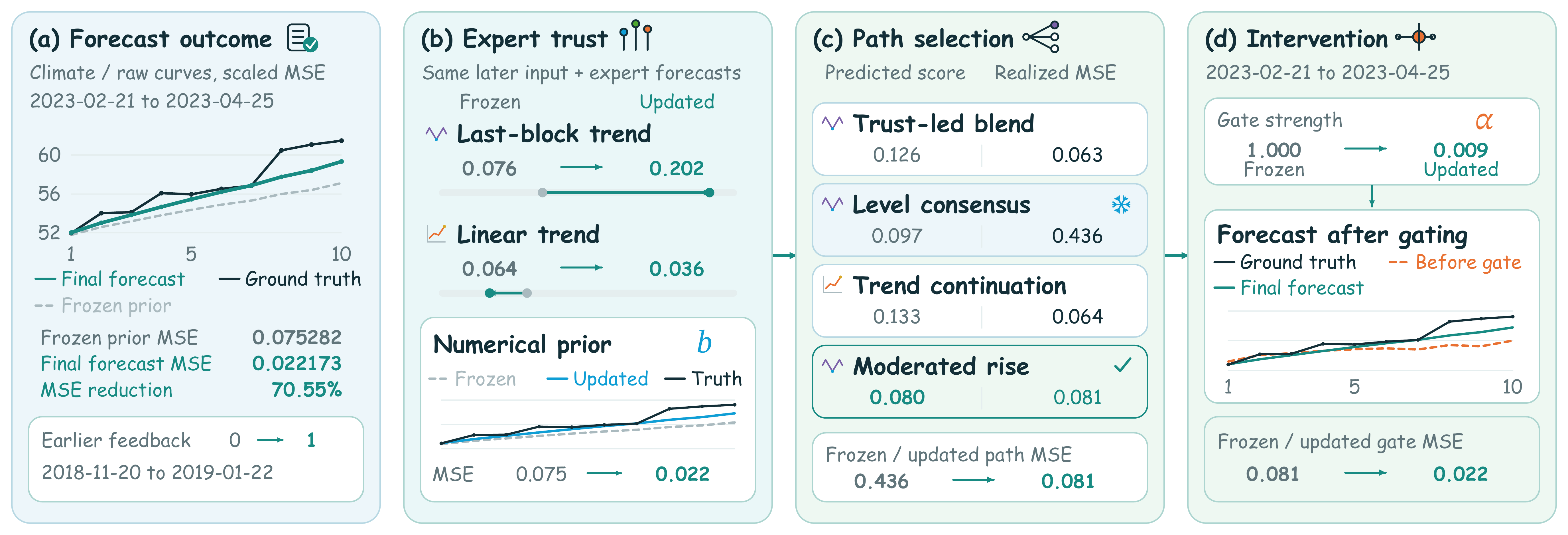}
    \caption{
    \textbf{Realized feedback reshapes future forecasting decisions.}
    A two-window Climate replay from Time-MMD using statistical forecasting experts.
    (a) Final forecast, ground truth, and Trust-frozen numerical prior.
    (b) Reweighting strengthens the prior's upward trend.
    (c) Path selection favors moderated rise over level consensus.
    (d) The learned gate limits premature flattening by the selected path.
    Panel (a) uses original units; all MSE values use training-split
    standardization.
    }
    \label{fig:case-study}
\end{figure*}

\subsection{Case study of policy evolution}
\label{sec:case-study}

To complement the aggregate comparisons, we inspect a post hoc selected
two-window Climate replay using the statistical-expert configuration of \method{} on Time-MMD~\citep{liu2024timemmd}.
The target continues rising and steepens toward the end of the horizon,
whereas the Trust-frozen prior responds too weakly to this movement
in \cref{fig:case-study}(a). The case examines how earlier feedback
revises later decisions. \appref{app:case-study} gives the sample dates,
replay protocol, and comparison settings.

\cref{fig:case-study}(b-d) reveals three complementary responses.
EvolveTrust shifts reliance toward the recent-block trend and away from
the full-window linear trend, producing a prior with stronger upward movement.
EvolveReason changes the preferred interpretation from level consensus
to a moderated rise, reducing the tendency to flatten the forecast.
The selected path nevertheless levels off earlier than the revised prior.
The intervention rule learned from earlier feedback assigns it limited
influence, preserving the prior's stronger continuation in the final forecast.

The mechanism therefore separates path preference from intervention
strength. Expert adaptation improves the numerical basis, path adaptation
changes the preferred interpretation, and the gate controls its effect
on the output. Their interaction illustrates how policy evolution
retains useful numerical structure while revising subsequent decisions.

\section{Conclusion}

\method{} turns realized futures into supervision for expert trust, agent path selection, and intervention strength through a temporally ordered predict, reveal, and update loop. Across eight Time-MMD domains, it achieves the best average MSE and MAE ranks among fifteen methods and improves both metrics over its fixed-policy counterpart in every domain. The case analysis illustrates how these updates change the interpretation and use of numerical forecasts. Together, the results support adapting the orchestration policy while keeping forecasting components frozen. The agent does more than retain past outcomes. It uses them to revise how it forecasts what comes next.

\section*{AI Use Statement}
We use generative AI tools to assist with manuscript drafting and polishing, literature retrieval and discovery, and figure preparation. We take responsibility for verifying all AI-assisted text, references, and figures and for the accuracy and integrity of the final content.

\section*{Ethics Statement}
As our work only focuses on the time series forecasting problem, there are no potential ethical risks.

\section*{Reproducibility Statement}
The main text defines the forecast construction and policy-update equations. \appref{app:timemmd-protocol} specifies dataset configurations, chronological splits, reveal eligibility, metrics, and aggregation. \appref{app:run-scope} describes the numerical tool pool, policy ablations, feedback restrictions, and case replay. Average ranks use all fifteen methods and mean ranks for ties at the reported precision.

\bibliography{iclr2027_conference}
\bibliographystyle{iclr2027_conference}

\clearpage
\appendix

\section{Limitations}
\label{app:limitations}

This study focuses on forecasting components and outcomes that are unaffected by forecast selection. Therefore, its predictive scope depends partly on the available base models. Richer expert sets and longer deployment studies are natural extensions of this approach. The predict, reveal and update protocol provides a basis for examining these settings while maintaining the temporal separation between prediction and feedback.

\section{Time-MMD Evaluation Protocol}
\label{app:timemmd-protocol}

Time-MMD aligns numerical targets with temporally indexed reports and web search text~\citep{liu2024timemmd}. The main comparison uses Agriculture, Climate, Economy, Energy, Environment, Security, Social Good, and Traffic, which are the eight domains shared with the KairosAgent evaluation.

The text modality separates report facts, report predictions, search facts, and search predictions. An item is available only when its recorded availability interval ends no later than the forecasting origin, and items overlapping the target window are excluded. \cref{tab:timemmd-domains} lists the domain-specific input lengths and forecast horizons used by the evaluation configuration. All lengths count observations, not calendar units. Climate retains the benchmark configuration of eight input observations and horizons $\{6,8,10,12\}$ even though its released numerical timestamps are weekly. This distinction also applies to the case study in \appref{app:case-study}. Energy and Environment use their separate weekly and daily configurations.

Each series is placed in chronological order, including Economy, before the $70/10/20$ training, validation, and test split. The scaler uses only observed training values. Filled numerical inputs preserve continuity, while an immutable validity mask excludes filled targets from every loss and policy update. Text availability is evaluated at each forecasting origin independently of the numerical sampling frequency.

We score every complete test origin with stride one. A forecast issued at origin $s$ becomes eligible for an update at origin $t$ only when $s+H\leq t$. For each domain, we evaluate the four configured horizons independently and average them with equal weight,
\begin{equation}
    \overline{m}_{d}=\frac{1}{|\mathcal{H}_d|}\sum_{h\in\mathcal{H}_d}m_{d,h},
\end{equation}
where $m_{d,h}$ denotes MSE or MAE for one domain and horizon~\citep{hu2026bridging,hu2025adaptive,timefilter}. We rank the fifteen methods within each domain using the reported three-decimal values, assign ties their mean rank, and average the resulting ranks across the eight domains.

\begin{table}[!ht]
\caption{Domain-specific forecast configurations. Cadence describes the numerical timestamps in the Time-MMD release; lookback and horizons count observations. Climate retains its benchmark shape configuration despite its weekly cadence.}
\label{tab:timemmd-domains}
\centering
\small
\begin{tabular}{llrl}
\toprule
Domain & Cadence & Lookback & Forecast horizons \\
\midrule
Agriculture & Monthly & 8 & $6,8,10,12$ \\
Climate & Weekly & 8 & $6,8,10,12$ \\
Economy & Monthly & 8 & $6,8,10,12$ \\
Energy & Weekly & 36 & $12,24,36,48$ \\
Environment & Daily & 96 & $48,96,192,336$ \\
Security & Monthly & 8 & $6,8,10,12$ \\
Social Good & Monthly & 8 & $6,8,10,12$ \\
Traffic & Monthly & 8 & $6,8,10,12$ \\
\bottomrule
\end{tabular}
\end{table}

\section{Implementation and Evaluation Details}
\label{app:run-scope}

\subsection{Evaluated configuration}

The numerical tool pool augments statistical experts with Chronos-2~\citep{ansari2025chronos2}, Toto-2.0 (313M)~\citep{khwaja2026toto2}, Falcon-X~\citep{liu2026falconx}, TimesFM-2.5~\citep{das2024timesfm}, and Chronos-Bolt~\citep{ansari2024chronos}. All numerical forecasters remain frozen. 

\method{} and \methodstack{} share the same foundation-model-augmented \pack{} forecasts, evidence and context pipeline, candidate-generation configuration, initialization, chronological origins, and aggregation rules. The complete method updates expert trust, agent path selection, and intervention strength after each eligible reveal. \methodstack{} holds all three policy components at their initial checkpoint throughout testing. This pairing isolates the effect of outcome-driven policy evolution while preserving the forecast and evidence substrate.

\subsection{Policy ablations and feedback coverage}
\label{app:ablation-settings}

The policy ablations in \cref{sec:policy-evolution-results} change which learned states may update. Freezing a component retains its inference-time computation and initial checkpoint; it does not remove numerical experts, candidate generation, or the final blending operation. Its outputs still depend on the current history and evidence. The feedback-coverage comparisons in \cref{sec:feedback-coverage-results} instead change which stored losses enter the update. A selected-path restriction still trains the selector, and a selected-expert restriction still updates trust. They therefore differ from freezing either learner.

\begin{table}[!ht]
\centering
\small
\caption{Update rules for the two families of comparisons. All and selected describe which committed alternatives supply loss feedback. Fixed means that the learned state remains at its initial checkpoint.}
\label{tab:ablation-settings}
\begin{tabular}{@{}llll@{}}
\toprule
Configuration & Expert feedback & Path feedback & Intervention \\
\midrule
\method{} & All & All & Updated \\
\methodstack{} & Fixed & Fixed & Fixed \\
w/o EvolveTrust & Fixed & All & Updated \\
w/o EvolveReason & All & Fixed & Updated \\
w/o EvolveIntervene & All & All & Fixed \\
Selected-path feedback & All & Selected path & Updated \\
Selected-expert feedback & Selected support & All & Updated \\
Selected-only feedback & Selected support & Selected path & Updated \\
\bottomrule
\end{tabular}
\end{table}

For a forecast issued at $s$, selected-expert feedback uses only $k\in S_s^{(j_s)}$, while selected-path feedback uses only $j_s$. These eligibility sets come from the pre-reveal record and remain fixed when the target arrives. In \cref{eq:online-trust}, an expert's eligible-origin set is restricted accordingly; in \cref{eq:reason-ridge}, the candidate sum is restricted to the recorded selected path. Intervention learning retains the blend target formed from the committed prior and selected candidate in every updating coverage regime. All-alternative feedback and the frozen-policy reference correspond to \method{} and \methodstack{}, respectively, so their repeated rows denote shared configurations.

\subsection{Climate case replay}
\label{app:case-study}

\cref{sec:case-study} uses the weekly Climate target column in the Time-MMD numerical release, whose source is historical drought information from Drought.gov~\citep{liu2024timemmd}. The later numerical forecast context consists of eight observations from December 27, 2022 to February 14, 2023. Its ten targets span February 21 to April 25, 2023. The earlier window uses numerical context from September 25 to November 13, 2018 and targets from November 20, 2018 to January 22, 2019. Past-reliability diagnostics use only observations available before each origin. The replay reveals the earlier target window, updates the policy once, and applies the resulting checkpoint to the later window. No forecasts or policy updates occur in the intervening period. The case is selected after outcome evaluation to inspect the behavior of all three update components.

The replay retains its recorded configuration of thirteen statistical filters and four structured candidate paths, with additive corrections disabled. This configuration provides an interpretable illustration of the update mechanism, separate from the foundation-model-augmented pool used in the aggregate experiments. Comparisons reuse the recorded expert-support choices and recompute expert weights, candidate execution, selection, and blending under the relevant policy state. Expert forecasts, executed candidates, and comparison outputs are stored before each target reveal. Targets and predictions use the same training-split scaler for scoring; \cref{fig:case-study}(a) converts the curves back to original units.

The panels in \cref{fig:case-study} distinguish component outputs from the final forecast. Panel (a) compares the final forecast with a numerical prior whose Trust state remains at its initial checkpoint. Panel (b) isolates prior construction under frozen and updated trust using the same expert forecasts. Panel (c) compares path selection under a fixed updated trust state. Its predicted scores are log-loss estimates available before reveal, while realized MSE is computed afterward. Panel (d) holds the updated prior and selected path fixed and changes only the intervention state. The final forecast includes the gated path contribution and is therefore close to, but distinct from, the updated prior. These comparisons explain the forecast's construction at a selected origin, separately from the aggregate evaluation over test origins.

\begin{figure}[!htbp]
    \centering
    \includegraphics[width=\textwidth]{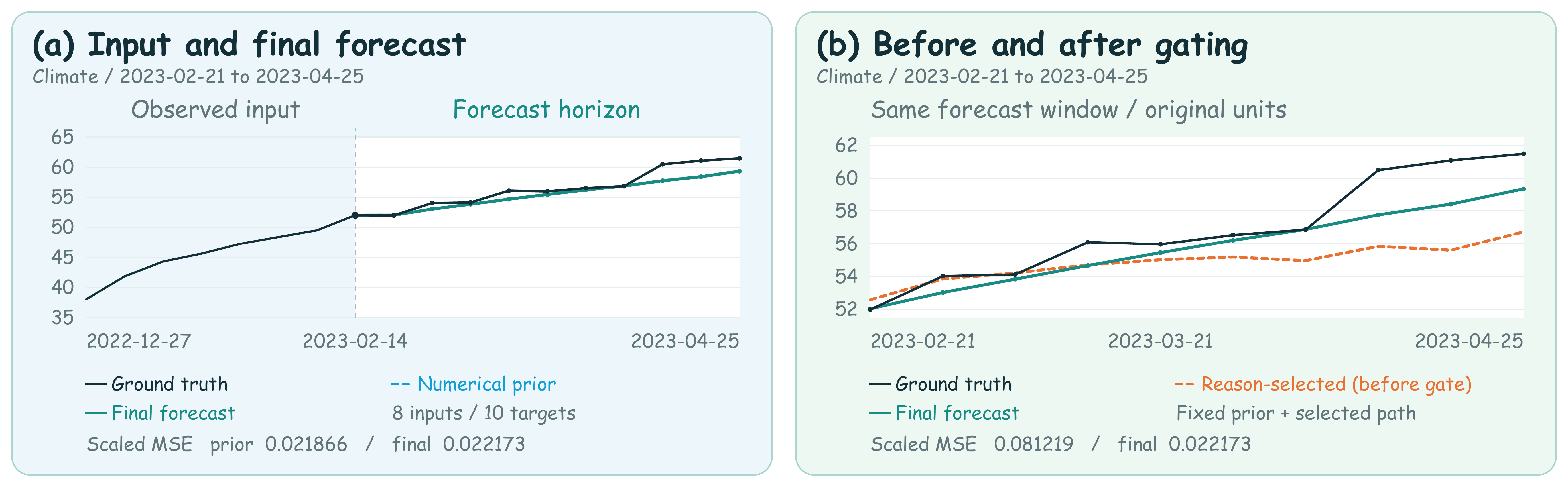}
    \caption{\textbf{Numerical input and forecast outputs for the Climate case.}
    (a) Eight observed inputs and the ten-target forecast window, together with the updated-trust numerical prior and the final forecast.
    (b) The selected candidate before gating and the final forecast over the same horizon.
    The numerical prior in panel (a) uses updated trust, unlike the frozen-trust reference in \cref{fig:case-study}(a).
    Curves use original target units with panel-specific vertical ranges; the displayed MSE values use training-split standardization.
    }
    \label{fig:climate-case-input-output}
\end{figure}

\cref{fig:climate-case-input-output} places these decisions in their temporal context. The observed history rises steadily, while the realized forecast window contains short plateaus followed by renewed growth and a sharper increase near its end. The selected candidate before gating follows the early level but becomes too flat to capture the subsequent rise. The final forecast stays close to the updated numerical prior, retaining its rising trajectory while attenuating the flatter candidate. It still underestimates the late increase. This behavior illustrates intervention as selective control of a candidate's influence, rather than an unconditional override of the numerical forecast.

\section{Method Details and Intervention Analysis}
\label{app:method-details}

\subsection{Evidence and candidate execution}

The \pack{} interfaces with frozen time series foundation models and statistical experts for level, moving average, robust trend, seasonality, spectral structure, state space, and shifts. The main evaluation uses the foundation-model-augmented configuration described in \appref{app:run-scope}, while the Climate replay retains its recorded statistical-filter configuration (\appref{app:case-study}). Each active expert receives the available numerical history and returns an aligned forecast, and unavailable members are excluded from the active set and its normalization.

Expert evidence includes historical RMSE/MAE, bias, exponentially weighted error, trend and seasonal residuals, boundary continuity, forecast range and volatility, source-relative consensus, and disagreement. A fixed tool bundle measures level, range, recent and global trend, volatility, mean absolute change, periodicity, autocorrelation, level shifts, and extremes. Context retains the distinction between report facts, report predictions, search facts, and search predictions. Items enter the context when their recorded availability intervals end by the forecasting origin, and target-overlapping items are excluded.

Typed validation precedes entry into the language-model context. Path validation rejects empty expert selections, zero trust mass, and invalid corrections. The source-budget projection retains a probability distribution on the selected subset while constraining source concentration. Its configured budgets and the admissible correction bounds determine the execution map; the language model supplies the structured choices, not unconstrained mixture coefficients. A local correction $\boldsymbol{\Delta}_t^{(j)}$ can be zero even when the overall deviation $\boldsymbol{\delta}_t=\va_t-\vb_t$ is nonzero because the path selects a different expert mixture.

\subsection{Rolling fits and stored state}
\label{app:rolling-fits}

Let $\mathcal{W}_t^{\mathrm{R}}$ and $\mathcal{W}_t^{\mathrm{I}}$ be the rolling sets of completed origins retained for path selection and intervention. Each retained origin satisfies $s+H\leq t$ and $m_s>0$, and intervention additionally requires identifiable correction energy. The ridge objectives corresponding to the two learners are
\begin{align}
    \boldsymbol{\theta}_t
    &=\arg\min_{\boldsymbol{\theta}}
    \sum_{s\in\mathcal{W}_t^{\mathrm{R}}}\sum_{j\in\mathcal{J}_s}
    \left({\boldsymbol{\psi}_s^{(j)}}^\top\boldsymbol{\theta}-z_s^{(j)}\right)^2
    +\boldsymbol{\theta}^\top\Lambda_{\mathrm{R}}\boldsymbol{\theta},
    \label{eq:reason-ridge}\\
    \boldsymbol{\omega}_t
    &=\arg\min_{\boldsymbol{\omega}}
    \sum_{s\in\mathcal{W}_t^{\mathrm{I}}}\bar w_s
    \left(\vg_s^\top\boldsymbol{\omega}-\alpha_s^\star\right)^2
    +\boldsymbol{\omega}^\top\Lambda_{\mathrm{I}}\boldsymbol{\omega}.
    \label{eq:intervention-ridge}
\end{align}
The positive-semidefinite matrices $\Lambda_{\mathrm{R}}$ and $\Lambda_{\mathrm{I}}$ encode the configured ridge penalties, including intercept treatment. The feature maps, rolling-window settings, cold-start conditions, and penalty choices are calibration settings. Each component state contains its coefficients and the buffers or sufficient statistics needed to update them. The active checkpoint is replaced once validation of all three component updates completes; if validation fails, the current checkpoint is retained. Retaining original features pairs each outcome with the decisions available before it was observed.

\subsection{Geometry of the intervention target}
\label{app:blend-derivation}

Fix a committed prior and selected path, with $m_t>0$ and $D_t>0$. Expanding \cref{eq:blend-risk} gives
\begin{equation}
    L_t(\alpha)=\frac{\|\vr_t\|_{\Omega_t}^{2}
    -2\alpha\langle\boldsymbol{\delta}_t,\vr_t\rangle_{\Omega_t}
    +\alpha^2D_t}{m_t}.
\end{equation}
Its derivative vanishes at $\tilde\alpha_t=\langle\boldsymbol{\delta}_t,\vr_t\rangle_{\Omega_t}/D_t$, and its second derivative is $2w_t>0$. This establishes both the unique unconstrained optimum and the clipped optimum in \cref{eq:alpha-star}. Completing the square gives \cref{eq:blend-quadratic}. For any admissible $\alpha\in[0,1]$, the exact excess loss is
\begin{equation}
\begin{split}
    L_t(\alpha)-L_t(\alpha_t^\star)
    &=w_t(\alpha-\alpha_t^\star)^2
    +2w_t(\alpha-\alpha_t^\star)(\alpha_t^\star-\tilde\alpha_t)\\
    &\geq w_t(\alpha-\alpha_t^\star)^2.
\end{split}
    \label{eq:boundary-excess}
\end{equation}
The second term vanishes for an unconstrained optimum inside $[0,1]$. If $\tilde\alpha_t<0$, both factors in that term are nonnegative; if $\tilde\alpha_t>1$, both are nonpositive. This explains the boundary correction. With $D_t=0$, the loss is constant in $\alpha$ on the observed coordinates, so that window contains no intervention-identification signal.

The regression in \cref{eq:intervention-ridge} predicts the clipped target using bounded energy weights. For any raw prediction $q$, projection satisfies $|\clip(q,0,1)-\alpha_t^\star|\leq|q-\alpha_t^\star|$. The ridge loss therefore provides a stable target-fitting objective for the bounded gate. It is a supervised surrogate rather than an unconditional identity with forecast MSE, because boundary projection, bounded weights, and parameter regularization change the optimization objective.

\end{document}